# Generating Multilingual Personalized Descriptions of Museum Exhibits – The M-PIRO Project

Ion Androutsopoulos[+], Vassiliki Kokkinaki[*], Aggeliki Dimitromanolaki[+],
Jo Calder[♣], Jon Oberlander[♣] and Elena Not[♠]

[+]Software and Knowledge Engineering Laboratory
Institute of Informatics and Telecommunications
National Centre for Scientific Research (NCSR) "Demokritos"
P.O. Box 60228, GR-153 10 Ag. Paraskevi, Athens, Greece
e-mail: {ionandr, adimit}@iit.demokritos.gr

[*]Foundation of the Hellenic World
38 Poulopoulou St., GR-118 51 Athens, Greece
e-mail: vkokk@fhw.gr

[♣]Division of Informatics, University of Edinburgh
2 Buccleuch Place, Edinburgh EH8 9LW, UK
e-mail: {J.Calder, J.Oberlander}@ed.ac.uk

[♠]Cognitive and Communication Technologies Division,
IRST, Istituto Trentino di Cultura,
via Sommarive 18 - Povo, 38050 Trento, Italy
e-mail: not@irst.itc.it

## Abstract

This paper provides an overall presentation of the M-PIRO project. M-PIRO is developing technology that will allow museums to generate automatically textual or spoken descriptions of exhibits for collections available over the Web or in virtual reality environments. The descriptions are generated in several languages from information in a language-independent database and small fragments of text, and they can be tailored according to the backgrounds of the users, their ages, and their previous interaction with the system. An authoring tool allows museum curators to update the system's database and to control the language and content of the resulting descriptions. Although the project is still in progress, a Web-based demonstrator that supports English, Greek and Italian is already available, and it is used throughout the paper to highlight the capabilities of the emerging technology.

## 1. Introduction

In recent decades, there has been significant effort in museums, galleries, and other "memory institutions" to make their collections more accessible to the public. A main target is to advance beyond museum tags, by providing more information in a coherent and pleasant way that treats the exhibits not only as interesting material objects, but also as carriers of valuable information for the understanding of the past

[Dallas 1993]. Collections available in electronic form, for example via the Internet or on CDs, are also increasingly common, and can be accessed by very wide audiences, augmenting the reach of traditional forms of presentations. As the audience of the collections grows, however, so does the need for mechanisms to tailor the available information to different target groups, including visitors speaking different languages, of different ages, and with different backgrounds and levels of expertise.

At the physical premises of museums and galleries, visitors can be accompanied by curators or professional guides, who use their experience to present information which is both interesting to the particular visitors and important from an educational point of view, often in the visitors' native languages. When visitors interact with electronic collections, however, curators and guides cannot usually be present, and information is more difficult to tailor. While it is possible, for example, to prepare in advance written or recorded descriptions of the exhibits for various ages or levels of expertise, this requires additional effort and increases the maintenance cost; and the cost increases significantly if the descriptions are to be translated into many languages. Furthermore, in settings where electronic visitors are allowed to view the exhibits in any order and revisit the collections several times, static pre-written descriptions or recordings are problematic, as they do not take into account what the visitor has already seen, and how long ago. For example, suppose a visitor sees several pieces of work by some sculptor, and later approaches yet another statue by the same artist. Repetition of information about the sculptor will then be redundant, unless the lapse in time is significant. It would be more constructive to draw the visitor's attention to differences between the current and the previous statues, or to convey new information, for example, about the creation period of the exhibit. The use of analogies and comparisons enhances informal learning, and turns the museum visit into a more fruitful experience [Hooper-Greenhill 1994]. This calls for ways to store information in a language- and visitor-independent manner, rather than pre-written or pre-recorded texts, and for mechanisms to convert dynamically this information to personalized textual or spoken descriptions of exhibits.

The M-PIRO project aims to develop technology that will address these needs. Using natural language generation, user modeling, and speech synthesis techniques, M-PIRO is developing technology that will allow information about exhibits to be stored in a database and to be converted into high-quality textual or spoken personalized descriptions in several languages (currently, English, Greek, and Italian) for presentation over the Web or in virtual reality environments. It is important to note that the descriptions are not written or recorded in advance, but are generated automatically from small fragments of texts and database fields that contain symbolic codes representing, for example, particular statues, sculptors, styles, etc. The same database is used to generate descriptions in all the languages, reducing dramatically translation costs and guaranteeing consistency across languages. Furthermore, an authoring tool is being developed to enable museum curators to update the source information, and control the language and contents of the resulting descriptions. As the descriptions are generated dynamically, they can be tailored in various respects, including the facts they convey, the language style and vocabulary, the length of the descriptions, and the hyperlinks suggested for further explorations.

M-PIRO builds upon the ILEX natural language generation system [Oberlander et al. 1998], which was originally used to produce dynamically exhibit descriptions for a

Web-based electronic gallery of 20th century jewellery.[1] M-PIRO extends ILEX's technology by incorporating improved multilingual capabilities, high-quality speech output, authoring facilities, extended user modeling mechanisms, as well as a more modular core generation engine. Furthermore, M-PIRO incorporates and extends machinery from the HIPS project [Benelli et al. 1999; Not & Zancanaro 2000], which allows pre-existing texts to be merged more naturally with dynamically generated descriptions; this is useful in cases where a piece of information is too difficult to be generated dynamically. We shall return to these points in following sections. Although M-PIRO is still in progress, a large-scale demonstrator is already available, and it will be used throughout this paper to demonstrate the capabilities of the emerging technology.

The remainder of this paper provides a general overview of M-PIRO, attempting to avoid technical details and focusing on the functionality that the project provides to curators and other museum staff. Section 2 presents the overall system architecture. Section 3 highlights the process that generates dynamically exhibit descriptions from the database. Section 4 presents the authoring tool that is under development. Section 5 concludes and discusses future work plans.

## 2. System architecture

Figure 1 shows M-PIRO's system architecture in abstract terms. In Web-based environments, visitors select exhibits to be described by clicking on thumbnail images, as shown in Figure 2, which is a screenshot from the current M-PIRO demonstrator. (The user interfaces shown in this paper are interim versions, which will be improved in later stages of the project.) In virtual reality presentations, exhibits are selected by approaching them. Once an exhibit has been selected, the system retrieves from the database all the information that is relevant to the exhibit, and using natural language generation [McDonald 2000; McKeown 1995; Reiter & Dale 1997, 2000] and speech synthesis techniques [Dutoit 1997] it produces an appropriate textual or spoken description. Figure 3 shows an English description generated by the current demonstrator, and Figure 4 shows the same description in Greek. The structure of the database and the generation process will be discussed further in Section 3.

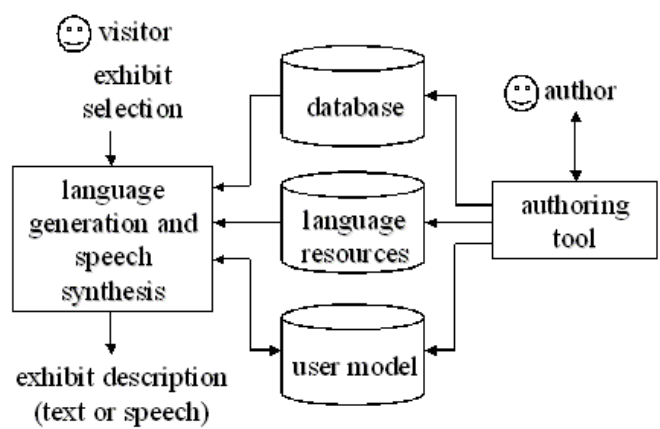

**Figure 1: M-PIRO's system architecture**

---
[1] See http://www.cstr.ed.ac.uk/cgi-bin/ilex.cgi .

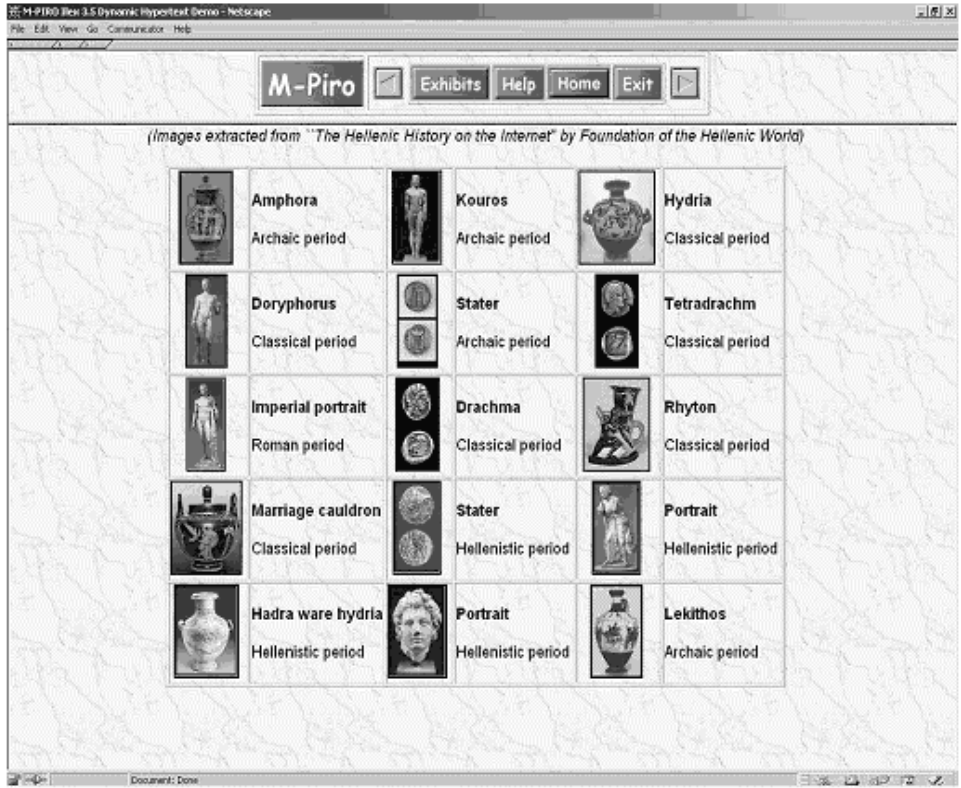

Figure 2: Selecting an exhibit in the Web-based demonstrator

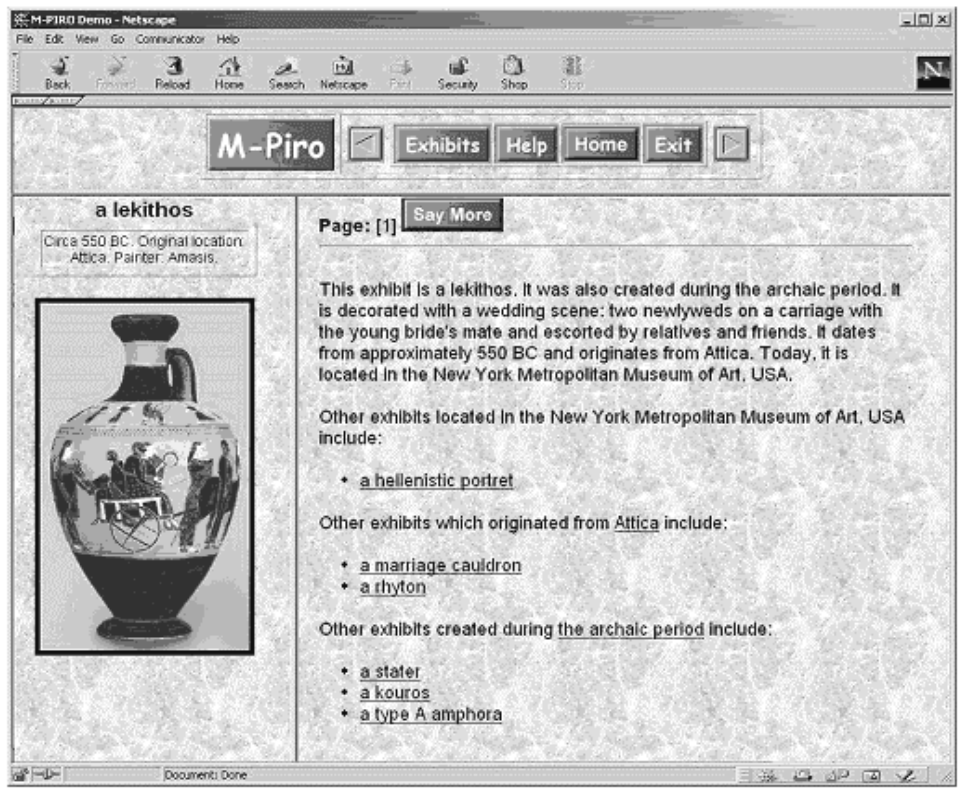

Figure 3: A dynamically generated exhibit description in English

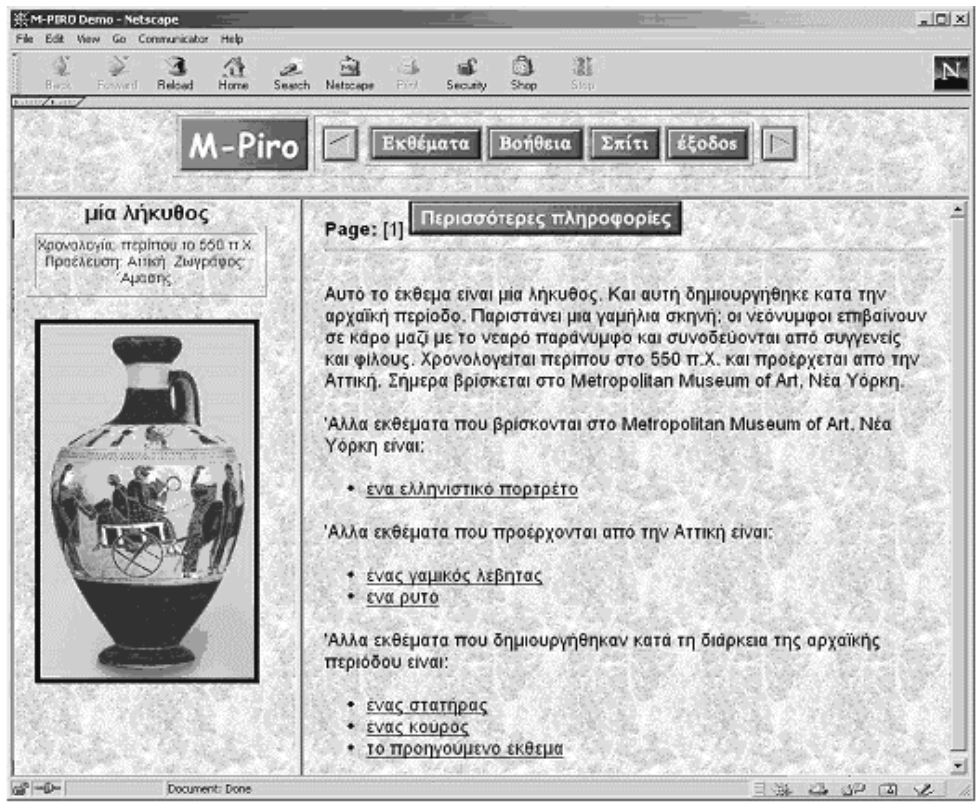

**Figure 4: A dynamically generated exhibit description in Greek**

The generation and synthesis components employ a variety of language resources; for example, lexicons and grammars of the supported languages, rules specifying which words can be used to express each concept, etc. Most of these resources are domain-independent; i.e., the system can be used with exhibits of different kinds, for example frescoes or jewels rather than statues, without modifying them. Some of the language resources, however, are domain-dependent, and need to be tuned to each collection domain, as part of the authoring process, to be discussed in Section 4.

A user model is also consulted during the generation of the descriptions. This is a collection of user preferences, such as the preferred language and length of exhibit descriptions, and records showing which exhibits the visitor has seen and what the system has told the visitor about them. These records allow the system to avoid repeating information that has already been conveyed, and to compare the current exhibit to previous ones [Milosavljevic & Dale 1997; Milosavljevic & Oberlander 1998; Milosavljevic 1999]. In Figure 3, for example, the description reminds the user that both the current and the previous exhibit were created in the archaic period, helping the visitor build a more coherent view of the collection. Following ILEX, the user model also contains scores indicating the educational value of each piece of information, as well as how likely it is for users of a particular type to find the information interesting [Mellish et al. 1998a; Oberlander et al. 1998]. For each exhibit, the database typically contains more information than can be expressed in a description of reasonable length. The system attempts to convey only facts that have not been expressed in the past, and among those, it focuses on facts of high interest and educational value. (This process will be discussed in more detail in Section 3.) The "say more" button (cf. Figure 3) allows the visitor to receive more information about the selected exhibit, until all the relevant facts in the database have been exhausted. This adheres to the principle that all the available information should ultimately be available to all the visitors. Additional user modeling mechanisms are being developed in M-PIRO to allow the system to adjust its vocabulary and language complexity according to the user type. With children, for example, it may be preferable to use shorter sentences and more common words (e.g., "shows" rather than "depicts") compared to texts for adults, and some domain experts may prefer the telegraphic style of traditional museum labels, rather than full text.

One of M-PIRO's most ambitious goals is that domain experts, such as museum curators, should be able to update the system's domain-dependent knowledge without the intervention of language technology experts. That is, it will be possible for domain experts, called *authors*, to configure M-PIRO for a new collection, and to inspect or modify the system's knowledge about a collection via an authoring tool (Figure 1), to be discussed in Section 4. Although some training and familiarity with computers will still be required, this constitutes a major advance compared to most natural language generation systems, where porting the system to a new domain requires programming and language technology expertise.

## 3. Dynamic generation of exhibit descriptions

Let us now examine more closely the process of generating exhibit descriptions from the database, starting from the structure and contents of the database. An entity-relationship database model is assumed; i.e., the database is taken to hold information about entities (e.g., statues, artists) and relationships between entities (e.g., the artist of each statue). Entities can be both concrete and abstract objects (e.g., historical periods or styles), and they are organized in a hierarchy of entity types. In the domain of M-PIRO's current demonstrator, for example, the basic entity types include – among others – the types "exhibit", "historical-period", "place", and "person". The "exhibit" entity type is further subdivided into "vessel", "statue", and "coin", as shown in the left pane of Figure 5, and "statue" is further divided into "kouros" and "imperial-portraits". Each entity is declared to belong to a particular entity type; for example, "exhibit 2" in Figure 5 is declared to be a "kouros" and, therefore, also a "statue" and an "exhibit", while "archaic-period" is a "historical-period". To make the system easier to use, we have opted for a single-inheritance hierarchy; i.e, each entity type can have only one direct super-type, and, excluding inheritance, each entity can belong to only one entity type. In fact, the core system can also handle multiple inheritance, but multiple-inheritance is not available to the users of the authoring tool, for the sake of simplicity.

Relationships between entities are expressed using fields. At each entity type, it is possible to introduce new fields, which then become available to all the entities of the type and its subtypes. For example, the "statue" type in Figure 5 introduces the field "sculpted-by"; consequently, all the entities of this type, including entities of type "kouros" and "imperial-portrait", will carry this field. The "creation-period" field is inherited from the "exhibit" type, and is, therefore, also available with non-statue exhibits, i.e., entities of type "vessel" and "coin".

The fillers of each field must be entities of a particular entity type. In Figure 5, the fillers of "sculpted-by" are declared to be entities of the type "sculptor", while the fillers of "creation-period" are required to belong to the type "historical-period". The latter allows entities like "archaic-period" and "classical-period" to be used as values of "creation-period". The "set-valued" option in Figure 5 allows a field to be filled by multiple fillers of the specified type; in the "previous-locations" field, this allows us to enter more than one previous locations of the exhibit.

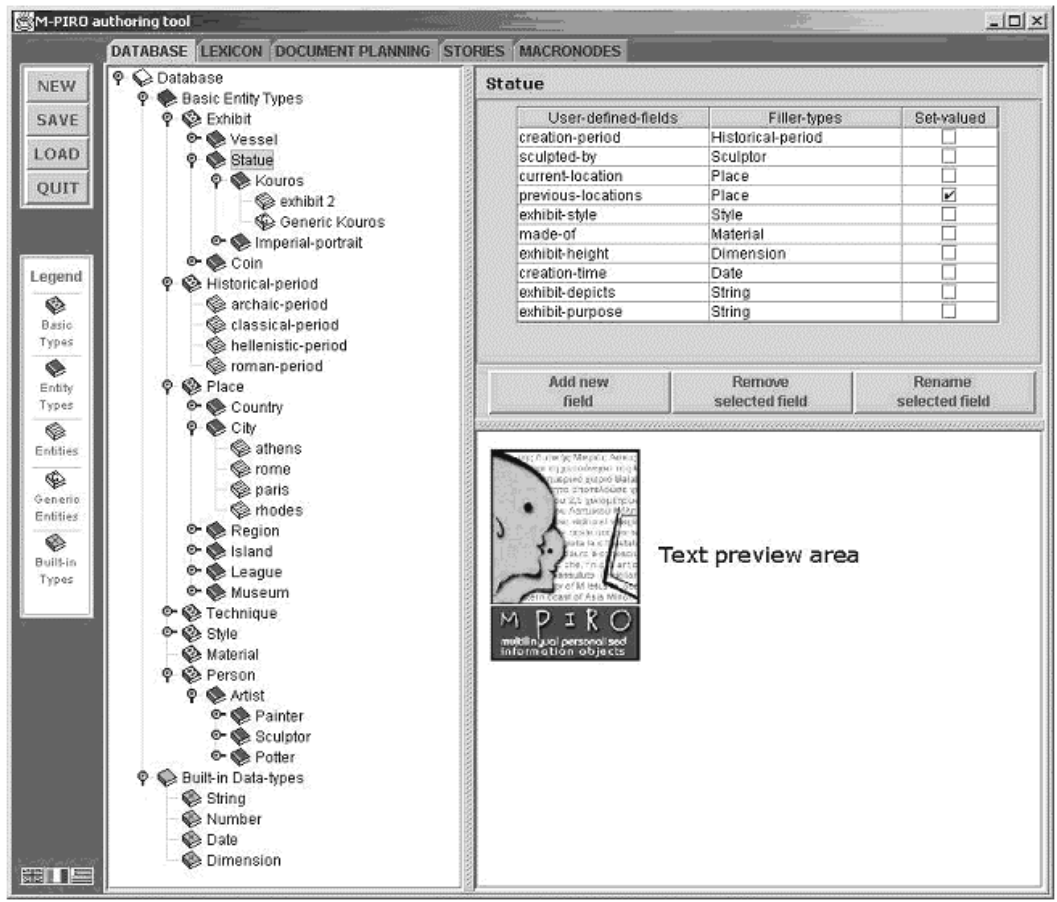

**Figure 5: Editing the entity types with the authoring tool**

Fields are also used to express attributes of entities, for example, their names or their dimensions. A number of built-in data-types are available, like "string" and "date", and these are used to specify the allowed values of attribute-denoting fields. In Figure 5, the "exhibit-depicts" and "exhibit-purpose" fields are declared to be string-valued attributes. They are intended to hold short canned texts in the three languages, describing what the exhibits depict and their purposes, as this information is typically too difficult to express using automatic text generation. Larger, paragraph-long canned texts can be associated with particular entities or entity types via the "stories" tab of Figure 5. In Figure 3, the sentence "It is decorated with a wedding scene: two newlyweds... and friends" is a canned string stored as the value of "exhibit-depicts"; the rest of the text is generated dynamically.

Following ILEX, M-PIRO associates with each fact in the database an *interest*, an *importance*, and an *assimilation* score (not shown in Figure 5) per user type [Mellish et al. 1998a; Oberlander et al. 1998]. The interest score shows how likely it is for a visitor of a particular type to find the fact interesting. Domain experts, for example, may be interested to see references to published articles that discuss the selected exhibit, while casual visitors would probably find such information uninteresting. The importance score, on the other hand, shows how important it is for the museum to convey the fact to each visitor type, the idea being that some uninteresting facts may, nonetheless, be important to convey. Finally, the assimilation score shows the extent to which the fact can be assumed to be known to the user, either from general knowledge or through previous interaction with the system. After being set by the author to an initial base value, the assimilation score changes dynamically during the interaction to help the system avoid repeating information the user already knows. The three scores are used during the first stage of the generation process, called *content selection*, to select the facts (relations or attributes) that the exhibit description should convey (Figure 6). Ideally, the facts must all be related to the selected exhibit, they must not have been expressed, they must be both interesting and educationally important, and they must not exceed a maximum number of facts per description [O'Donnell 1997]. Future Web-based M-PIRO demonstrators will allow the visitor to adjust the maximum number of facts interactively, effectively allowing the visitor to select the desired length of the descriptions.

The next stage in the generation process is *document planning* (Figure 6). This outputs the overall document structure, which specifies, for example, the desired sequence of the facts in the generated description and their rhetorical relations; for example, whether a fact amplifies or contrasts another one [Hovy 1993; Mann & Thompson 1988]. M-PIRO has inherited from ILEX a variety of domain-independent document planners [Mellish et al. 1998a,b], which are being extended to allow the authors to specify domain-dependent schema-like planning rules [McKeown 1995] to capture the particular structural characteristics of museum descriptions. Museum labels, for example, typically start with information about the type and creation period of the exhibits. The curator of a collection of coins may wish to specify that descriptions should then proceed with a description of what the two sides of the coin depict, followed by information about the material and style, with bibliographical references given a much lower priority. In many descriptions, they might not be presented at all. While some aspects of content selection and document planning may in general be language-dependent, in the context of museum collections we have so far managed to produce good-quality descriptions in all three languages (English, Greek, and Italian) using ILEX's content selection and document planning processes, which were originally developed for English. In contrast, the next two stages, *micro-planning* and *surface realization*, are language-specific, and separate resources had to be developed for each language.

Micro-planning specifies in abstract terms how a fact can be expressed as a phrase in each language; for example, which verb to use, in what voice and tense, and which participating entities of a relationship should be expressed as subject and object. In the case of the "sculpted-by" relationship of Figure 5, micro-planning could specify that the verb "to sculpt" should be used in English, and that the verb should be rendered in passive past form, with the subject expressing the statue and the agent expressing the sculptor. This would give rise to sentences like "This statue was sculpted by Polyklitus". Micro-planning specifications of this form are provided for each database field and language, using authoring facilities that will not be covered in this paper. Ongoing work investigates how multiple micro-planning specifications per field can be exploited (e.g., allowing both "to sculpt" and "to create" to be used in "sculpted-by" in both passive and active forms) to allow for greater variety in the resulting descriptions, and to tailor the vocabulary and style of the descriptions according to the visitor type. Micro-planning also includes processing steps that determine which facts can be aggregated in a single sentence (e.g., "This vase dates from approximately 550 BC and was found in Attica" rather than "This vase dates from approximately 550 BC. It was found in Attica."), and what type of referring expression should be generated for each entity (e.g., "Doryphorus", "this statue", or "it" may be more or less appropriate in the context of previous sentences); these steps normally do not require any input from the authors, and will not be discussed further.

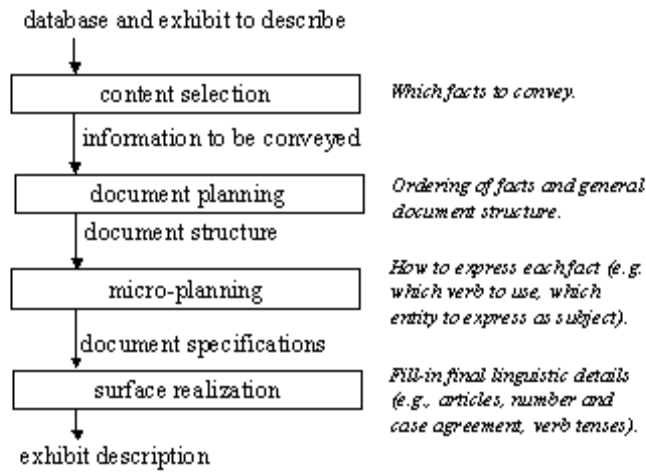

Figure 6: Stages of natural language generation in M-PIRO

The last generation stage, *surface realization*, is responsible for producing the final textual form of the descriptions. This includes producing the appropriate word forms (e.g., verb tenses) based on the document specifications output by micro-planning, placing the various constituents (e.g., subject, verb, object, adverbials) in the correct order, accounting for number and gender agreement, etc. Surface realization is based on large-scale systemic grammars [Halliday 1985; Teich 1999], one for each supported language, that capture the necessary linguistic information. While the grammars are domain-independent, the lexicons that they employ are to a large extent domain-specific, and they need to be tuned when the system is ported to a new domain; this will be discussed further in Section 4. M-PIRO employs ILEX's grammar of English, which is in turn based on the WAG system [O'Donnell 1994]. The Greek grammar was constructed by taking the English grammar as a starting point, and gradually modifying the elements of the grammar where the two languages differ [Bateman et al. 1991; Bateman 1997; Dimitromanolaki et al. 2001; Kruijff et al. 2000]. The Italian grammar was developed in a similar manner, starting from a grammar for Spanish [Morales-Cueto 1998].

Once a textual description has been generated, it can also be passed to a speech synthesizer to produce a spoken form. This is particularly useful in virtual reality tours, where dynamic visual content may make textual descriptions impractical. M-PIRO is developing high-quality synthesizers for all three languages, aiming to exploit additional speech-related markup that can be added to the textual descriptions. Unlike many `text-to-speech' applications, the fact that the texts are generated by computer means that significant additional information can be provided to the synthesizer. Factors such as phrasal boundaries, rhetorical relations between phrases, and whether some item of information has been previously expressed can then be taken into account in producing synthesized speech [Theune et al. 2001]. This approach is expected to lead to improved prosody, adding to the acceptability of the system in real-usage scenarios.

## 4. Interactive symbolic authoring

Let us now examine more closely M-PIRO's authoring tool (see Figure 1). The tool is intended to be used at two stages: *domain authoring* and *exhibit authoring*. Domain authoring provides the general information about the domain, which includes the available entity types, their fields, and domain-dependent language resources, while exhibit authoring allows particular entities (e.g., exhibits, artists) to be entered into the database and their fields to be filled. Although the authoring tool is designed to be used by domain experts with no language technology expertise, some general training on the use of the tool is still required at both stages. Most of the routine authoring in a museum is expected to concern exhibit authoring, which is easier to master. It is, thus, possible to train only a few museum staff members as domain authors; they will be responsible for the initial configuration of the system and "advanced" modifications, such as the addition of new types of exhibits, or the tailoring of the domain-dependent linguistic mechanisms. The task of entering and maintaining the information about individual exhibits can then be assigned to a larger number of exhibit authors, who may have received briefer training on the use of the authoring tool.

Domain authoring starts with the construction of the hierarchy of entity types and the definition of the fields that will be available at each type, as discussed in Section 3. For each entity type, the domain author then specifies one or more nouns that can be used to refer to the entities of the type; for example, the noun "statue" and its Italian and Greek equivalents "statua" and "άγαλμα", respectively, can be used to refer to entities of type "statue". The authoring tool forces the vocabularies of the supported languages to be kept aligned, guaranteeing that equivalent texts can be generated in all languages. Morphological components are also provided, that save the author from having to enter all the forms of each noun. This is particularly important for highly inflected languages like Italian and Greek, as demonstrated in Figure 7, where the system has generated automatically all the forms of the Greek noun "άγαλμα" from its base form. Each entity type inherits the nouns that have been associated with its super-types, and thus nouns need only be associated with the most general types they can be used with. In a similar manner, facilities are provided that allow the domain author to specify which verbs can be used to express each relationship or attribute, as well as the tense and voice of the resulting phrase, and other micro-planning specifications (Section 3).

A final stage of domain authoring, shown as *document planning* in the top menu of the authoring tool in Figure 7, is the specification of the document structure, which includes the desired sequence of the facts in the generated descriptions and their rhetorical relations (Section 3). This aspect of authoring will not be discussed further in this paper, as related work in M-PIRO is still in progress. It is expected that future versions of the authoring tool will also allow the domain author to specify user types (e.g., child, casual adult visitor, archaeologist) and assign different values to user modeling parameters, such as the interest and importance scores of the facts, depending on the user type.

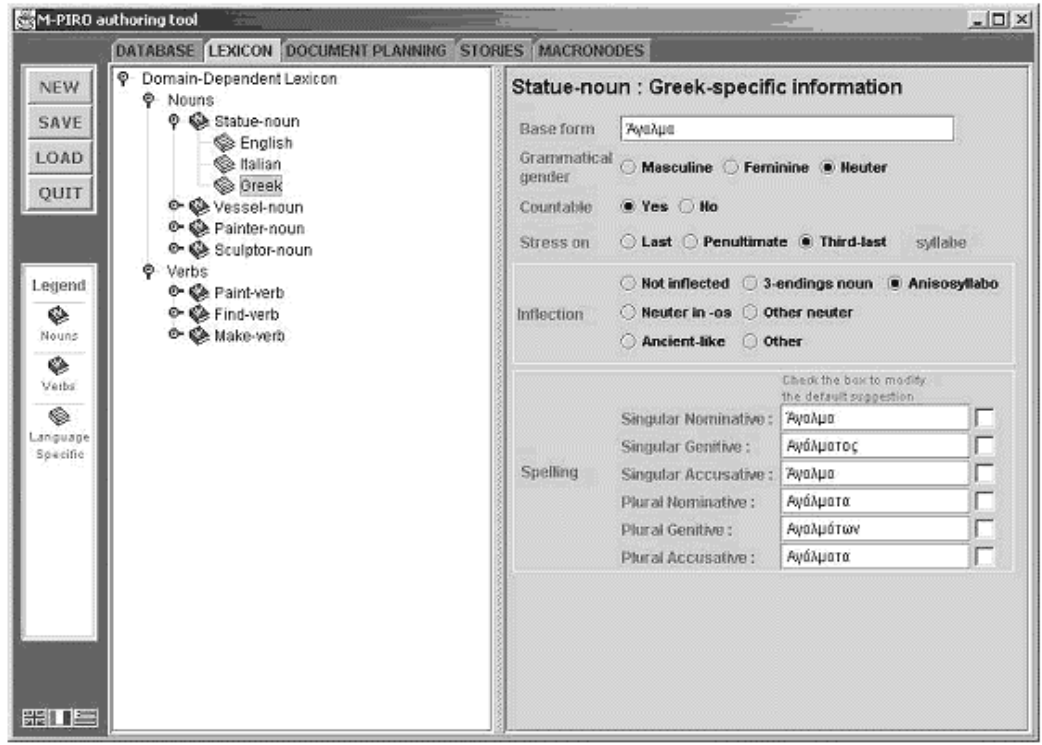

**Figure 7: Updating the domain-dependent lexicon**

Once the general information about the domain has been provided, exhibit authors can start entering information about the exhibits. This amounts to inserting entities in the appropriate entity types, and filling in their slots with appropriate symbolic codes. This is illustrated in Figure 8, where information about a kouros has been inserted (cf. Figure 5). To capture default information about all the entities of a type, *generic entities* can be introduced. For example, the "creation-period" field of the "generic kouros" entity in Figure 8 could be assigned the value "archaic-period". This would indicate that, unless otherwise mentioned, a kouros belongs to the archaic period, saving the exhibit author from having to specify this information for each individual kouros.

Domain authoring in practice overlaps with the beginning of exhibit authoring. That is, domain authors typically have to insert a few exhibits of each type before they can decide about the final form of the hierarchy of entity types, their fields, and the micro-planning and document structure specifications. This is an incremental process, whereby initial information about the domain is used to generate preliminary descriptions of sample exhibits, with the generated descriptions then guiding the domain author to elaborate further the domain information. For example, new fields may be necessary to express information that is missing, new words may be needed to express new fields, and micro-planning or document structure specifications may need to be adjusted to generate texts that sound more natural. This process is repeated until the domain author is satisfied with the structure and content of the sample descriptions, at which point exhibit authors may start entering larger numbers of exhibits.

The authoring tool provides interactive facilities, designed to speed up the domain authoring process and help monitor the quality of the generated descriptions. M-PIRO expands previous ideas on interactive authoring [Paris et al. 1995; van Deemter 2000] by allowing the author to experience the effect of modifying not only the database entries, but also the database structure and the domain-specific linguistic resources. At any point, the author can ask the system to generate a description of a particular exhibit, based on the information that is currently in the database, using the current version of the domain-specific linguistic resources. This is illustrated in Figure 8, where a preview of the English description of an exhibit has been generated in the bottom right pane from the information in the database. Previews are available in all three supported languages, and it is also possible to preview phrases expressing particular fields of the database.[2] Furthermore, future versions of the authoring tool are expected to provide previews of different lengths and for different user types.

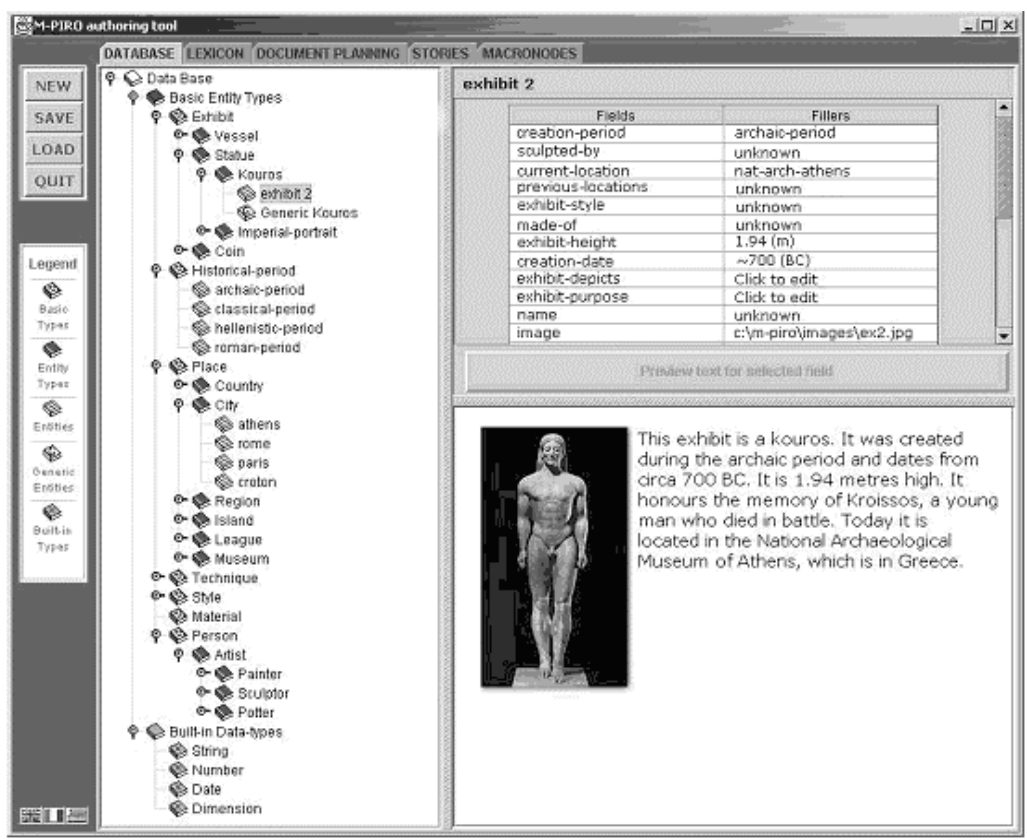

**Figure 8: Entering and previewing information about an exhibit**

## 5. Conclusions and future work

M-PIRO's technology will allow museums and other memory institutions to enhance their electronic collections by adding textual and spoken exhibit descriptions that are generated automatically. The descriptions are produced in several languages from a single database, which reduces dramatically translation costs, and they can be tailored according to what the visitors have already seen, their backgrounds, and language skills. M-PIRO's authoring tool allows domain experts, such as museum curators, to configure the technology for new collections, to inspect and modify the system's knowledge, and to control and monitor the content and linguistic form of the resulting texts. A large-scale Web-based prototype already demonstrates the benefits of the emerging technology.

---

[2] The authoring tool is currently under development, and previews of this form can only be generated off-line. Figure 8 demonstrates the previewing facilities that are being developed.

In subsequent stages of the project, M-PIRO's technology will be ported to a virtual reality environment, where visitors can approach and examine exhibits in three-dimensional space. In an immersive environment of this type, the ability to produce *spoken* descriptions is crucial, as it is very unnatural to require visitors to read texts. Further work is also planned to allow authors to tailor the overall structure of the generated descriptions, and to allow database records to be imported from existing museum databases. Another strand of ongoing work is developing methods to allow natural language generation to be combined with *macronodes*, a technology deriving from the HIPS project [Not and Zancanaro 2000] that allows canned texts to be customized according to the user's model, providing many of the benefits of full generation. The evaluation of M-PIRO's technology is another major component of future work. The functionality of the authoring tool will be evaluated with the help of museum experts, while the efficiency of the overall system in information provision and interaction with its visitors will be evaluated through formative and summative methods [Keene 1998]. Finally, it should be noted that apart from museum collections, M-PIRO's technology can be exploited in a variety of other contexts, including educational software, video games and on-line catalogues for electronic commerce.

## Acknowledgements

M-PIRO is a project of the Information Societies Programme of the European Union, running from February 2000 to January 2003. The project's consortium consists of the University of Edinburgh (UK, co-ordinator), ITC-irst (Italy), NCSR "Demokritos" (Greece), the University of Athens (Greece), the Foundation of the Hellenic World (Greece), and System Simulation Ltd (UK). More information about the project is available from: http://www.ltg.ed.ac.uk/mpiro/. The authors feel particularly indebted to Mick O'Donnell, formerly with ITC-irst, who contributed significantly to the project during its first year.